\documentclass[11pt]{article}

\usepackage[preprint]{acl}

\usepackage{times}
\usepackage{latexsym}

\usepackage[T1]{fontenc}

\usepackage[utf8]{inputenc}

\usepackage{microtype}

\usepackage{inconsolata}

\usepackage{graphicx}

\usepackage{hyperref}
\usepackage{url}
\usepackage{booktabs}
\usepackage{amsmath}
\usepackage{multirow}
\usepackage{multicol}
\usepackage[table]{xcolor}
\usepackage{subcaption}
\usepackage{caption}
\usepackage{svg}
\usepackage{listings}
\usepackage{amsmath,amssymb,amsthm}

\usepackage{adjustbox}
\usepackage{wrapfig}
\definecolor{rlcmshade}{RGB}{242,247,253}
\definecolor{promptshade}{RGB}{248,249,251}
\definecolor{promptrule}{RGB}{210,215,222}
\definecolor{takeawaybg}{RGB}{239,252,251}
\definecolor{takeawayrule}{RGB}{170,230,225}
\lstdefinestyle{promptblockstyle}{
    basicstyle=\ttfamily\small,
    breaklines=true,
    columns=fullflexible,
    keepspaces=true,
    frame=single,
    framerule=0.3pt,
    rulecolor=\color{promptrule},
    backgroundcolor=\color{promptshade},
    xleftmargin=0.5em,
    xrightmargin=0.5em,
    aboveskip=0.6em,
    belowskip=0.6em
}
\lstnewenvironment{promptblock}
    {\lstset{style=promptblockstyle}}
    {}
\newsavebox{\takeawayboxsave}
\newenvironment{takeawaybox}
    {\par\medskip\noindent\begin{center}\begin{lrbox}{\takeawayboxsave}\begin{minipage}{0.92\linewidth}}
    {\end{minipage}\end{lrbox}\setlength{\fboxsep}{7pt}\fcolorbox{takeawayrule}{takeawaybg}{\usebox{\takeawayboxsave}}\end{center}\medskip}

\title{Rethinking LoRA Memory Through the Lens of KV Cache Compression}

\author{
  Chunsheng Zuo \quad Liaoyaqi Wang \quad William Jurayj \quad  William Fleshman \quad Benjamin Van Durme \\
  Johns Hopkins University \\
  \texttt{\{czuo3,lwang240,wjurayj1,will.fleshman,bvandur1\}@jhu.edu}
}

\begin{document}
\maketitle

\begin{abstract}
Parametric retrieval augmentation encodes document information into lightweight, document-specific modules such as LoRA adapters, reducing the need to include all evidence as input context. However, it remains unclear how this parameter-side memory interacts with context-side memory stored in the KV cache. We study this interaction in document-level question answering by progressively evicting document key-value states and measuring when a document LoRA contributes beyond the retained context. We find that document LoRA adds little when the KV cache is largely intact, but becomes increasingly useful under aggressive compression, recovering 13–21 ROUGE-L points when no document context remains. 
The gain is largest when the base model encodes the document, and the adapter is applied only during answer generation, suggesting that document LoRA is better understood as decoding-time parametric memory than as a document encoder.
Finally, QA-style supervision produces substantially stronger adapters than raw-context next-token-prediction. These results position document LoRA as a complementary memory channel whose value emerges precisely when context-side evidence is scarce.
\end{abstract}

\section{Introduction}

\begin{figure*}[h]
    \centering
    \includegraphics[
        width=\textwidth,
        trim=0pt 5pt 0pt 0pt,
        clip
    ]{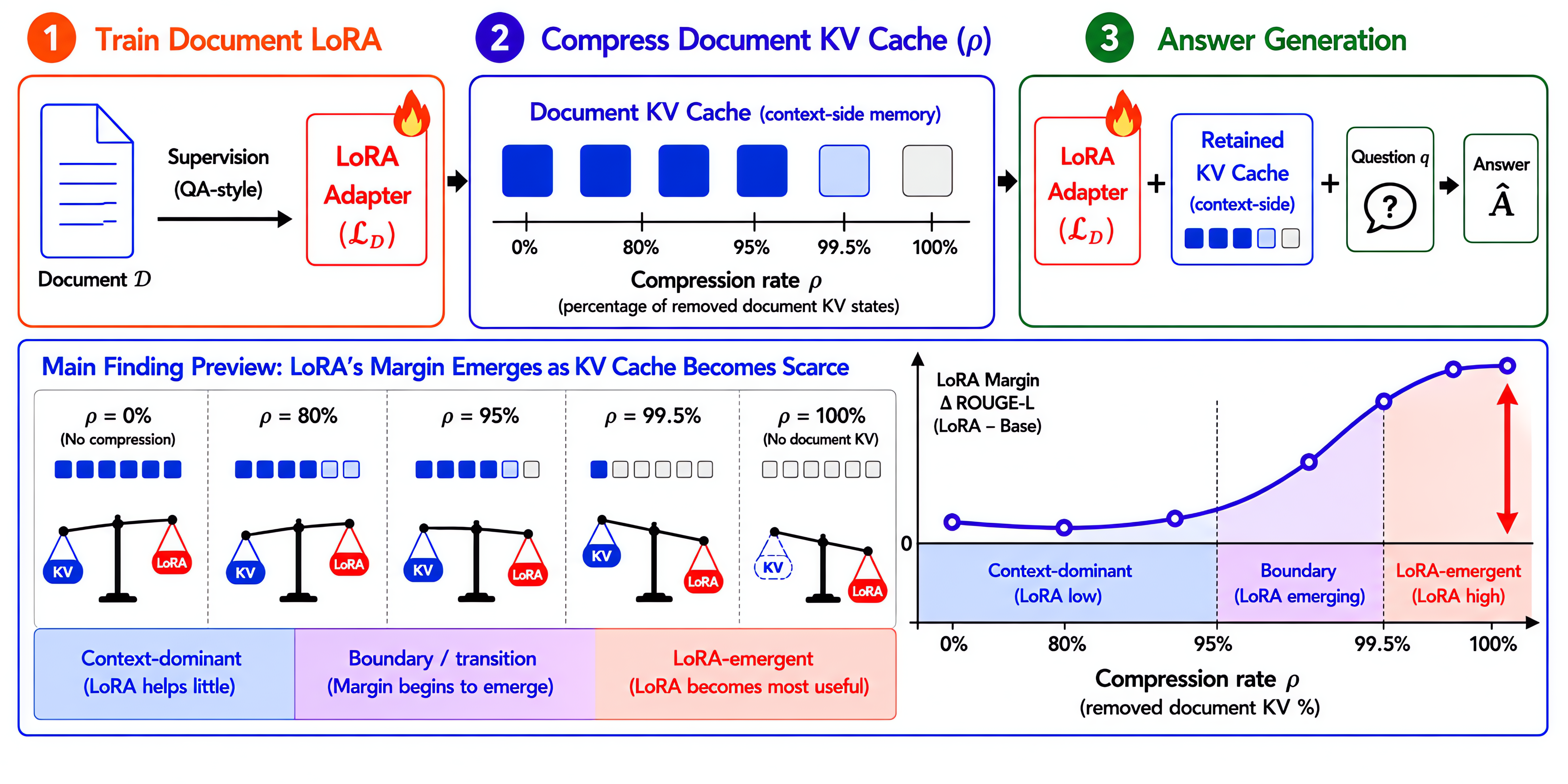}
    \caption{
   \textbf{Pipeline overview (Top).} We train a document-specific LoRA adapter with QA-style supervision (Step 1), compress the document KV cache by a factor $\rho$ (Step 2), and generate the answer using the compressed cache together with the adapter (Step 3). \textbf{Main findings (Bottom).} Varying $\rho$ traces out three regimes: at low compression the KV cache dominates and LoRA contributes little (context-dominant); under aggressive compression the LoRA margin ($\Delta$ ROUGE-L over base) begins to emerge (transition); and under extreme compression, the adapter becomes the dominant document-memory channel and yields the largest gain (LoRA-emergent).
    }
    \label{fig:teaser}
\end{figure*}

Retrieval-augmented generation typically provides external knowledge by appending retrieved documents to the input context~\citep{lewis2020rag}. While this gives the model direct token-level access to evidence, each retrieved token incurs prefill computation and contributes key-value states that must be stored throughout decoding, making long or repeatedly queried documents expensive.

This has motivated two lines of work to reduce dependence on having full documents in the KV cache. The first is KV-cache compression, especially the KV-cache eviction method, which drops less important tokens while keeping the potentially more useful ones inside the KV cache to save memory while preserving most performance~\citep{zhang2023h2o,liu2023scissorhands,li2024snapkv,chari2025compactor}. Another line of work instead explores parametric retrieval augmentation: encoding document information into lightweight document-specific parameter modules, often instantiated as LoRA adapters, which are loaded and used during inference~\citep{hu2021lora,su2025parametric,caccia2025km,back2026understanding}. This direction reframes external knowledge as memory carried by model parameters. These lines of work are complementary, but they are usually studied in isolation, making their interaction underexplored. 

In this work, we study this interaction in document-level question answering. For each document, we train a document-specific LoRA adapter from supervision derived from the document, and evaluate it while progressively compressing the document portion of the KV cache. This protocol, illustrated in \autoref{fig:teaser}, lets us compare two sources of document information: explicit evidence retained in the KV cache and the memory carried by the adapter. We further separate the stages at which LoRA is applied during inference: prefill, compression scoring, and decoding, and compare different ways of converting the document into adapter supervision.

Our experiments lead to three main findings:
\begin{itemize}
    \item \textbf{Document LoRA is most useful in the high-compression regime.}
    Its benefit is limited when the KV cache retains sufficient evidence, but it increases as compression removes that evidence.

    \item \textbf{Document LoRA is more useful as a decoding-time memory than as a document-prefill module.}
    The base model is better suited to construct the compressed document KV cache, while document LoRA is more useful during answer generation.
    
    \item \textbf{QA-style supervision produces the strongest document LoRAs for QA.}
    It outperforms other document-derived training objectives in our current training-format comparison.
\end{itemize}

\section{Related Work}

\subsection{KV Cache Compression}
The KV cache grows linearly with sequence length and batch size, often surpassing the model parameters in memory footprint and becoming the dominant cost in long-context serving~\citep{liu-etal-2024-lost}. KV-cache compression methods include \emph{eviction}, which keeps only KV entries selected by attention scores~\citep{zhang2023h2o,liu2023scissorhands,li2024snapkv}, positional heuristics~\citep{xiao2023streamingllm}, or learned budgets~\citep{yang-etal-2024-pyramidinfer,feng2024adakv}; \emph{merging}, which consolidates less important entries to mitigate eviction loss~\citep{zhang2024cam,wang2024kvmerger,wan2024d2o}; and \emph{quantization}, which reduces the bit-width of stored states~\citep{liu2024kivi,hooper2024kvquant,yang2024mikv}. These categories are not mutually exclusive: eviction and merging change which states remain or how they are combined, while quantization changes the precision of retained states.
In this work, we use KV-cache compression as a controlled way to vary how much document evidence remains available in context.

\subsection{Parameter-Side Document Memory}
Another line of work reduces reliance on long retrieved contexts by moving document information into trainable parameter modules, often implemented as LoRA adapters. These methods inject retrieved knowledge as low-rank weight updates rather than prompt tokens to avoid context bloat~\citep{liu-etal-2024-lost} and knowledge conflicts~\citep{xu-etal-2024-knowledge-conflicts}. Per-document adapters can be obtained offline via QA-pair fine-tuning~\citep{su2025parametric} or context distillation~\citep{caccia2025km}, or generated on the fly by a hypernetwork mapping document embeddings to LoRA weights~\citep{tan2025dyprag}. When a library of adapters already exists, LAG~\citep{fleshman2025lora} routes among them per token and per layer without further training.

Closer to our setting, empirical analyses~\citep{back2026understanding,tang2025parametric} characterize LoRA's capacity as parametric memory and show that parametric and in-context injection are largely complementary, while recent adapter-serving work studies delayed or selective LoRA activation to reuse base-model KV caches~\citep{greenewald2026activated,li2025efficient}. 
These studies, however, either study LoRA under full-context or no-context settings or treat stage-wise LoRA activation as an efficiency mechanism. We instead use KV-cache compression as a controlled axis that varies how much context-side document memory is retained, and use inference-stage controls as a diagnostic to ask where along this axis document LoRA contributes and which stages should apply it.

\section{Methodology}

\label{sec:design}
We study document-specialized LoRA as a form of parameter-side document memory under various KV cache compression settings. For each document, we train a separate LoRA using supervision derived from that document, then evaluate it on questions over the same document under different amounts of retained document KV cache. 

This section describes the training procedure (\autoref{sec:design-document-lora}), the KV-cache compression protocol (\autoref{sec:design-compression}), the evaluation protocol (\autoref{sec:evaluation-protocol}), and the inference variants used to separate LoRA's roles across stages (\autoref{sec:design-modes}) and supervision formats (\autoref{sec:design-training-format}).

\subsection{Document-Specialized LoRA Training}
\label{sec:design-document-lora}

To internalize a document into parametric memory, we follow a similar procedure as prior works~\cite{su2025parametric,caccia2025km,back2026understanding}. For each document $d$, we split the text into chunks
$\mathcal{C}_d=\{c_i\}_{i=1}^{n_d}$ and derive supervised training examples
from those chunks. A supervision format $f$ maps the chunks to a training set
$\mathcal{D}_d^{(f)}$. Each example $(x,m)\in \mathcal{D}_d^{(f)}$ contains a
token sequence $x$ and a loss mask sequence $m\in\{0,1\}^{|x|}$ that has the same length as $x$ to indicate which token the loss will be calculated on. Let $\theta_0$ be the frozen base model.
We train the document LoRA parameters $\Delta\theta_d^{(f)}$ by
\begin{equation}
\arg\min_{\Delta\theta}
\sum_{(x,m)\in \mathcal{D}_d^{(f)}}
\sum_{t=1}^{|x|}
-m_t
\log p_{\theta_0+\Delta\theta}(x_t \mid x_{<t}).
\end{equation}
Only the LoRA parameters are updated. We refer to the resulting adapter as the document LoRA $L_d$. At evaluation time, questions from document $d$ are answered with its document LoRA $L_d$.

\subsection{KV-Cache Compression}
\label{sec:design-compression}

Our compression method is based on Compactor~\citep{chari2025compactor}, a recent KV eviction algorithm that is training-free and robust (see verification in \autoref{sec:compression-method-robustness}). We use it to compress the document portion of the KV cache at a fixed rate. For a given layer and attention head, let the document key-value states be $\mathbf{K},\mathbf{V}\in\mathbb{R}^{N\times d}$, where $N$ is the number of document tokens. Compactor assigns each token an importance score by combining an attention-based score $a_i$ with an approximate leverage score $\widetilde{\ell}_i$:
\begin{equation}
    s_i =
    \frac{a_i-\bar{a}}{\operatorname{std}(\mathbf{a})}
    + \lambda
    \frac{\widetilde{\ell}_i-\overline{\widetilde{\ell}}}{\operatorname{std}(\widetilde{\boldsymbol{\ell}})} .
\end{equation}
Here $a_i$ is computed from non-causal attention mass, while $\widetilde{\ell}_i$ approximates the statistical leverage of the token's key vector. Let $\rho$ denote the compression rate, defined as the fraction of document key-value states removed, and let $r=1-\rho$ be the retention rate. We keep the indices
\begin{equation}
    \mathcal{I}_{\rho}
    =
    \operatorname{TopK}\left(\mathbf{s}, \left\lceil rN \right\rceil\right),
\end{equation}
and form the compressed document cache by retaining
$\mathbf{K}_{\mathcal{I}_{\rho}}$ and $\mathbf{V}_{\mathcal{I}_{\rho}}$. In our experiemnts, we evaluate $\rho\in\{0, 80\%, 90\%, 95\%, 99\%, 99.5\%, 100\% \}$, where $\rho=0$ keeps the full document KV cache, while $\rho=100\%$ drops the document entirely. The extreme compression ratios are chosen because we found they can be reflective of LoRA's effect. Implementation details can be found in \autoref{app:compression-implementation}.

\subsection{Evaluation protocol}
\label{sec:evaluation-protocol}
At evaluation time, each document comes with several questions. For each document, we first encode the user's conversational prefix followed by the document into the KV cache. For KV cache eviction, we only do it over the document portion of the KV cache, keeping the user's instruction prefix unchanged. After compression, we encode the question with the model and the compressed cache, then continue generating the answer by greedy decoding.
Appendix~\ref{app:qa-prompt-formats} details the prompt templates.

\subsection{Inference-Stage Controls}
\label{sec:design-modes}

Recent adapter-serving work uses selective LoRA activation to reuse base-model KV caches and reduce inference cost~\citep{greenewald2026activated,li2025efficient}. We repurpose stage-specific LoRA activation as a diagnostic: does document LoRA help more when constructing the document KV cache, or when decoding from a compressed one?
To answer this, we separate inference into three stages (document prefill, compression scoring, and answer decoding) and define four modes that vary LoRA usage across them. 
Let $B$ denote running the frozen base model without the adapter, and let $L_d$ denote running the same base model with the LoRA of document $d$ enabled. A cell in the table indicates whether that stage uses the base-only computation or the document-LoRA-augmented computation. The four modes specify how LoRA usage is assigned across the 3 inference stages.

\begin{center}
\small
\resizebox{\linewidth}{!}{
\begin{tabular}{@{}lccc@{}}
\toprule
Mode & Prefill/KV & Score & Decode \\
\midrule
Base & $B$ & $B$ & $B$ \\
Adapter score + adapter prefill & $L_d$ & $L_d$ & $L_d$ \\
Base KV + adapter decode & $B$ & $B$ & $L_d$ \\
Base score + adapter prefill & $L_d$ & $B$ & $L_d$ \\
\bottomrule
\end{tabular}
}
\end{center}

\paragraph{Base.}
The base model performs document prefill, compression scoring, and answer decoding without any adapter.

\paragraph{Adapter score + adapter prefill (default).}
The document LoRA is active throughout inference: it produces the document KV states, provides the compression scores, and decodes the answer.

\paragraph{Base KV + adapter decode.}
The base model performs document prefill and compression scoring, producing the compressed document KV cache. The document LoRA is then loaded only for answer decoding.

\paragraph{Base score + adapter prefill.}
The document LoRA produces the document KV states and decodes the answer, while compression scores are computed without the adapter. The retained indices from base scoring are then applied to the adapter-produced document KV states.

\subsection{Training-Format Controls}
\label{sec:design-training-format}

We compare four supervision formats derived from each document while keeping the document-specific training protocol fixed. Appendix~\ref{app:training-prompts} gives the exact prompt and loss-mask format for QA and raw-context supervision.

\paragraph{Synthetic QA (default).}
We prompt the model to generate QA pairs from document chunks. Each training sequence contains the generated question and answer, and the loss is applied only to answer tokens. This is the default format used in the main experiments unless otherwise stated. More details for synthesis are in \autoref{app:training-supervision-construction}

\paragraph{Raw Context.}
For each chunk, the training sequence is the document itself, and the loss is applied to all tokens of the document. 

\paragraph{Context Continuation.}
For adjacent chunks, we use the first chunk as context and the following chunk as the prediction target. The loss is applied only to the target chunk. 

\paragraph{Context + QA.}
This format includes the source context together with the generated question, while still applying the loss to the answer. 

\section{Experiment Setup}
\label{sec:design-data-train}

\paragraph{Datasets.}
We use NarrativeQA~\citep{kocisky-etal-2018-narrativeqa} and LongHealth~\citep{adams2024longhealth} as the main document-level QA testbeds. For NarrativeQA, each document corresponds to one story. For LongHealth, we treat one patient or case record as one document, so that the same per-document adapter protocol applies. Our evaluation subsets contain 20 documents each, with 596 QA examples for NarrativeQA and 400 QA examples for LongHealth. Using the Llama-3.1-8B-Instruct tokenizer, documents average
16K tokens for NarrativeQA and 11K tokens for LongHealth. Additional dataset statistics are given in Appendix~\ref{app:dataset-stats}.

\paragraph{Models.}
We use \texttt{Qwen3-4B}~\citep{yang2025qwen3technicalreport} and
\texttt{Llama-3.1-8B-Instruct}~\citep{grattafiori2024llama3herdmodels} as base models, both of which are instruction-finetuned models that can follow a user-assitant conversational format for question answering. These models also generate the synthetic data to train their own LoRA's.

\paragraph{Default LoRA recipe.}
Unless otherwise specified, we use document chunks of word length $64$ and train LoRAs with DoRA enabled~\cite{liu2024dora}, $r=16$, $\alpha=32$, and dropout rate $=0.1$ on the MLP projection modules. Adapters are trained for 4 epochs with learning rate $1\times10^{-4}$, batch size $4$, and weight decay $0.001$, using Synthetic QA supervision (\S\ref{sec:design-training-format}). 
All trainings are conducted on 1 H100 GPU with 96GB of memory.
Additional implementation details are in \autoref{sec::method details}.

\paragraph{Metrics.}
We report ROUGE-L as the main QA metric. Our QA setting uses generated phrase-level answers rather than fixed labels, so accuracy is not directly applicable, and exact match is too brittle to small wording or formatting differences. ROUGE-L gives a graded measure of answer overlap, which is better suited for comparing compression and adapter conditions.

\begin{figure*}[ht]
    \centering

    \includegraphics[width=\linewidth]{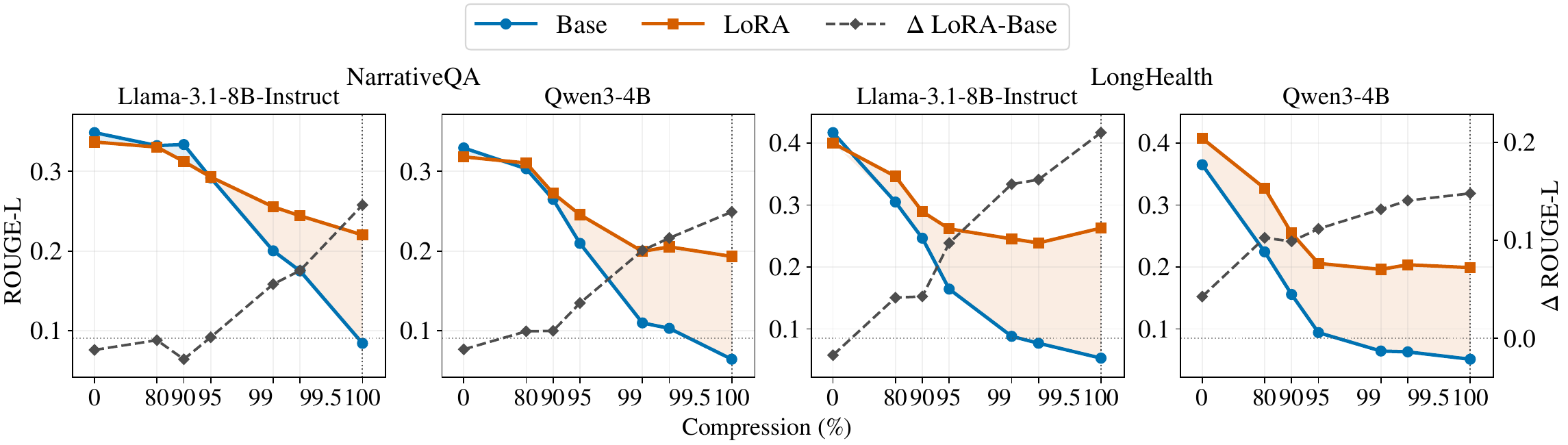}

    \caption{
    \textbf{Document LoRA complements aggressive KV-cache compression.}
    We compare QA performance with only the compressed document KV cache against performance with the corresponding document LoRA. Across \textbf{(a) NarrativeQA} and \textbf{(b) LongHealth}, the LoRA margin is small when much of the document KV cache remains, but grows under aggressive compression and at the no-context endpoint. 
    }
    \label{fig:main-compression-curve}
\end{figure*}

\section{Main Findings}

We treat document LoRA and the compressed KV cache as two complementary sources of document memory. In this section, we ask how they interact across compression regimes and organize the results around three diagnostic questions:
\begin{itemize}
    \item when does document LoRA become useful as document KV states are removed (\autoref{sec:lora-complements-compression});
    \item where should the adapter be active during inference: document prefill, compression scoring, or answer decoding (\autoref{sec:prefill});
    \item what supervision format makes adapter-side memory useful for QA (\autoref{sec:qa-only}).
\end{itemize}

\subsection{Document LoRA complements aggressive KV-cache compression}
\label{sec:lora-complements-compression}

\paragraph{LoRA adds little at low compression}
\autoref{fig:main-compression-curve} shows when document LoRA starts to matter. Across NarrativeQA and LongHealth, the same pattern emerges. When much of the document KV cache remains, the base model and the LoRA-augmented model remain relatively close because explicit document evidence is still available. On NarrativeQA, the LoRA margin is near zero or negative through most low-compression settings. LongHealth shows positive margins earlier, but these remain below the high-compression and no-context gains. 

\paragraph{LoRA becomes the primary document memory under extreme compression.}
The contrast becomes clearest at the high-compression end of the curve. At the no-context endpoint, the LoRA--base gain is roughly 13 ROUGE-L points on NarrativeQA and about 15--21 points on LongHealth. The margin curves, therefore, show that the adapter's contribution is not constant; it becomes visible precisely when context-side document memory is removed.

These results suggest a division of labor between the two memory sources. The KV cache remains the stronger carrier of document information when it is available, while document LoRA acts as a complementary parametric memory that partially recovers information lost under extreme compression. At 99\% compression, where only about 160 NarrativeQA tokens or 111 LongHealth tokens remain (\autoref{tab:remaining-context-tokens}), the adapter already recovers roughly half of its full no-context margin, suggesting that the transition from context-dominant to LoRA-emergent behavior occurs around this point.

\begin{takeawaybox}
\textbf{Implication:} Document LoRA should not be understood as a general replacement for long-context encoding. Instead, it is most useful as a memory augmentation mechanism under severe KV-cache budget constraints.
\end{takeawaybox}

\begin{figure*}[h]
    \centering
    \includegraphics[width=\textwidth]{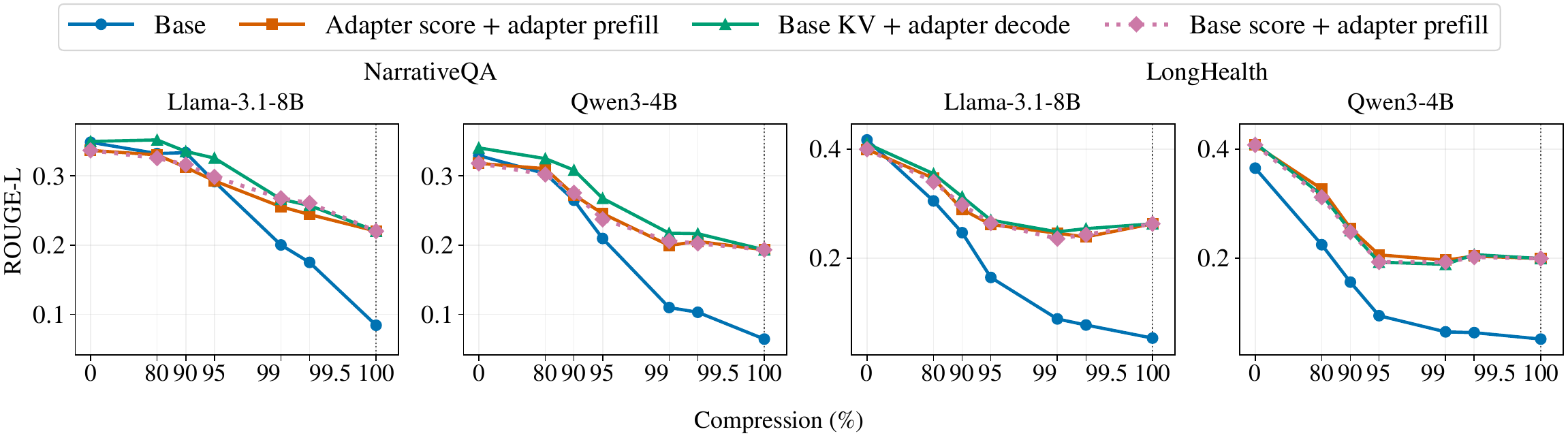}
    \caption{
    \textbf{Inference-stage controls.}
    We compare four ways of applying document LoRA across document prefill, compression scoring, and answer decoding on NarrativeQA and LongHealth. Base uses no adapter; Adapter score + adapter prefill enables LoRA throughout inference; Base KV + adapter decode uses the base model for document prefill and compression scoring, then enables LoRA for decoding; Base score + adapter prefill keeps LoRA-produced KV states but uses base-model compression scores.
    }
    \label{fig:prefill-diagnostic}
\end{figure*}

\subsection{QA-trained LoRA Helps More During Generation Than Document Encoding}
\label{sec:prefill}

We next ask where the adapter should be active once the document KV cache is compressed. Since document LoRA can affect the document KV states, the compression scores used to select retained states, and the final answer decoding, \autoref{fig:prefill-diagnostic} compares four inference modes that separate these 
roles.

\paragraph{Base-prefill with adapter decoding is strongest or competitive.}
At all compression points where some document KV remains ($\rho<100\%$), using the base model to prefill and compress the document KV cache and then enabling the document LoRA for decoding is usually the strongest or competitive mode. The pattern is clearest on NarrativeQA, where Base KV + adapter decode gives the best high-compression average for both models. It also holds for LongHealth with Llama-3.1-8B, while LongHealth with Qwen3-4B is closer to a tie and sometimes slightly favors using the adapter throughout inference. At the most severe non-empty compression points (99–99.5\%), Base KV + adapter decode improves over base-only inference by about 7–11 ROUGE-L points on NarrativeQA and 13–17 points on LongHealth. The question is therefore not whether the adapter helps, but where it should be applied: the strongest recipe is usually to let the base model build the document KV cache and reserve the adapter for generation.

\paragraph{The score-control condition does not make scoring the main explanation.}
The Base score + adapter prefill condition tests whether adapter-prefill is weaker simply because adapter-based compression scores select worse tokens. Changing only the scoring path has a small and inconsistent effect: it sometimes improves adapter-prefill, but it does not consistently match the base-prefill trajectory. This suggests that the central difference is not only which tokens are retained, but also which model produces the retained KV states. Base-produced document KV states provide stronger context-side memory, while the document LoRA is more useful as decoding-time parameter-side memory.

\begin{takeawaybox}
\textbf{Implication:} The stronger recipe is to use the base model for document prefill and apply the document LoRA primarily during answer generation. In this setting, LoRA is better understood as generation-time memory support rather than as a replacement document encoder.
\end{takeawaybox}

\subsection{QA-Style Supervision Produces the Strongest Document LoRAs}
\label{sec:qa-only}

\label{sec:training-format-results}
\begin{figure*}[ht]
    \centering
    \includegraphics[width=\linewidth]{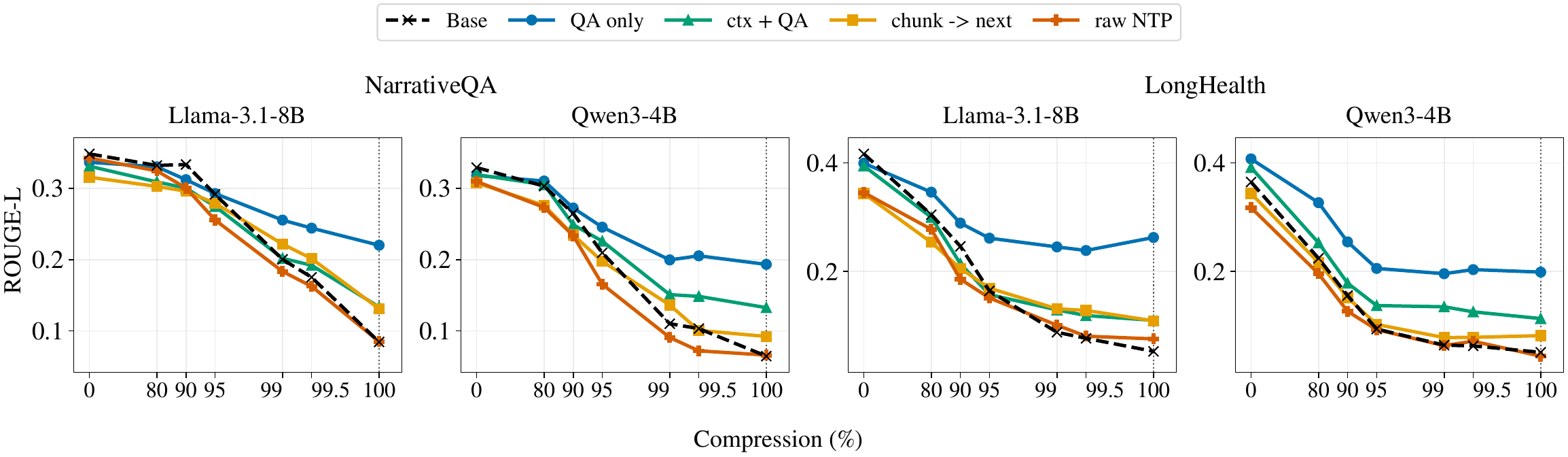}
    \caption{
    Training-format comparison on NarrativeQA and LongHealth with Llama-3.1-8B and Qwen3-4B. Each document LoRA is trained from the same documents using a different supervision format, then evaluated under KV-cache compression. QA-only supervision is strongest in the high-compression and no-context regimes across both datasets, while raw next-token prediction remains close to the base no-context behavior.
    }
    \label{fig:format-comparison}
\end{figure*}

We next vary the supervision used to train document LoRA while keeping the evaluation task fixed to document QA. \autoref{fig:format-comparison} compares QA-only supervision with three alternatives: raw next-token prediction on document text, chunk-to-next-chunk continuation, and context-plus-QA training where the source context is included in the training prompt. This comparison asks whether useful adapter-side memory comes from document exposure alone, from continuation-style training, or specifically from answer-supervised QA training.

\paragraph{QA-only supervision is strongest when context evidence is scarce.}
Across both datasets and both backbones, QA-only supervision produces the strongest adapter in the high-compression and no-context regimes. At the no-context endpoint, QA-only reaches about 0.19--0.26 ROUGE-L, while context-plus-QA is around 0.11--0.13, chunk continuation around 0.08--0.13, and raw next-token prediction remains close to the base no-context level. The separation is largest when the model can no longer rely on retained document KV states, showing that the training format matters most when the adapter has to provide the missing document signal.

\paragraph{Document exposure and QA formatting are not sufficient.}
Raw next-token prediction directly exposes the adapter to document text, but it does not produce strong QA behavior under compression. Chunk continuation is sometimes better than raw NTP at the most compressed points, but remains well below QA-only supervision. The context-plus-QA control is especially informative: it uses a QA interface, but because the source context is present during training, the model can answer from the prompt rather than pressure the adapter to store answer-bearing information. Thus, the advantage of QA-only supervision is not just document fine-tuning or QA-like formatting; it better matches the way the adapter must be used at evaluation time, where it needs to recall and express document information in response to questions.

\begin{takeawaybox}
\textbf{Implication:} A useful document LoRA cannot be obtained by simply reciting the document text. The adapter needs to learn to map the question to the answer, such that important facts can be recalled when context is absent during question answering.
\end{takeawaybox}

\section{Ablation Studies}
\label{sec:ablations}

In this section, we consider two design choices that directly shape the interaction between compressed context and document LoRA: \textbf{compression algorithm} and \textbf{LoRA target module group}. The compression algorithm determines which document KV states remain available, while the LoRA target module group determines where document-specific information can be stored in the model. We therefore ablate both choices.

\begin{figure*}[t]
    \centering
    \includegraphics[width=\linewidth]{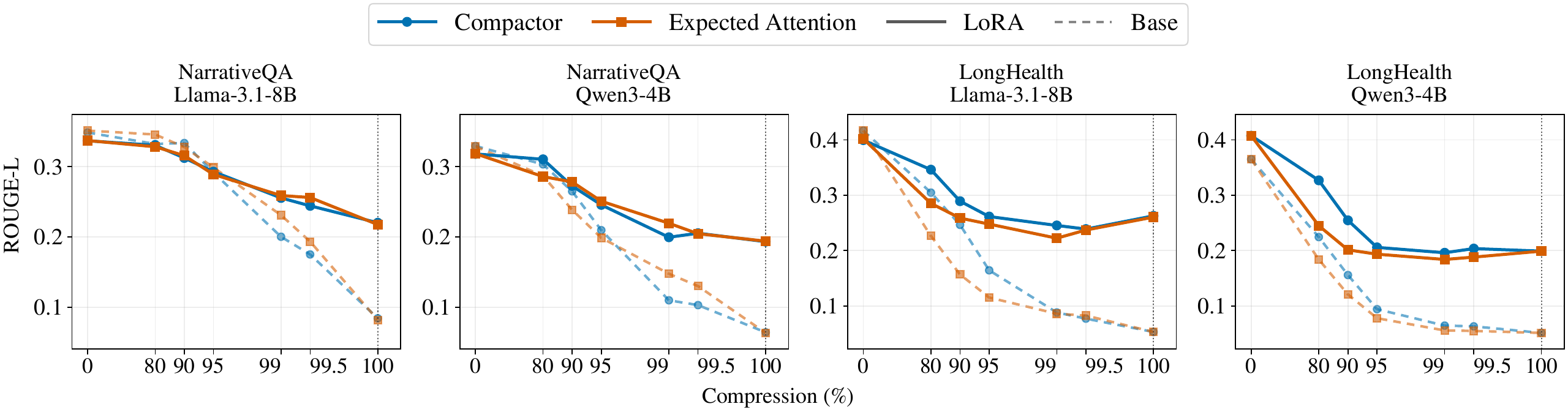}
    \caption{
        Compression method ablation. Each panel plots ROUGE-L against compression ratio for a given model, comparing base (no adapter) and document LoRA runs under two compression algorithms: Compactor and Expected Attention. \textbf{The margin between LoRA and base follows the same pattern for both algorithms: negligible at low compression ratios, and widening as the compression ratio increases.}
    }
    \label{fig:compression-method-ablation}
\end{figure*}

\begin{figure*}[t]
    \centering
    \includegraphics[width=\linewidth]{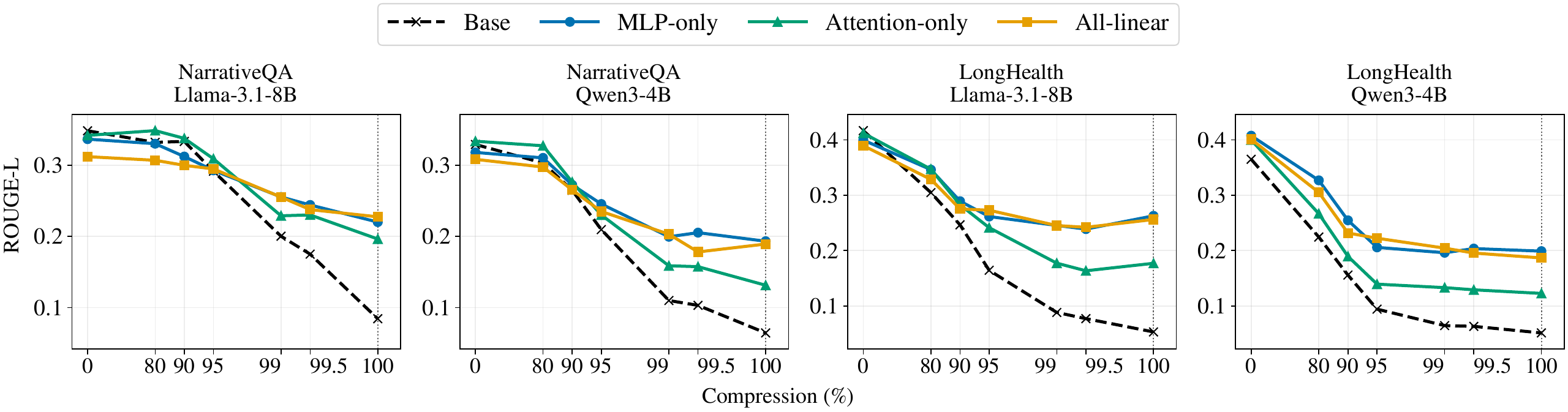}
    \caption{
        Target module ablation. Each panel plots ROUGE-L against compression ratio for three LoRA target configurations: attention-only, MLP-only, and all-linear, alongside the base model without any adapter. \textbf{MLP-targeted adapters are strongest or close to strongest at high compression ratios and in the no-context setting, while all-linear adaptation does not consistently improve over MLP-only.}
    }
    \label{fig:recipe-ablation}
    \label{fig:target-module-ablation}
\end{figure*}

\subsection{Compression Method Robustness}
\label{sec:compression-method-robustness}

Different KV-cache eviction rules can retain very different document tokens, so the compression-regime effect we observe with Compactor could in principle hinge on its specific scoring rule rather than on KV-cache compression in general. \autoref{fig:compression-method-ablation} tests this by replacing Compactor with Expected Attention~\citep{devoto2025expectedattention}, which ranks document KV entries by their average attention weight. The two compressors produce different absolute degradation curves: at 10\% retention on LongHealth/Llama, for instance, Expected Attention drops base ROUGE-L to 0.157 versus Compactor's 0.246. Crucially, however, the gap between LoRA and base follows the same pattern under both compressors---small margins when much of the document cache remains, and clear separation once compression becomes aggressive. This consistency across eviction rules supports the view that document LoRA compensates for missing document evidence rather than interacting with any particular compressor's selection bias.

\subsection{Target-module ablation}
\label{sec:target-module-ablation}

Prior interpretability and model-editing work suggests that transformer feed-forward modules are natural sites for storing and modifying semantic or factual associations: FFN layers can behave like key-value memories, and MLP weights are effective targets for factual edits~\citep{geva-etal-2021-transformer,dai-etal-2022-knowledge}. This motivates our default use of MLP-targeted document LoRA, but it does not by itself establish that MLP targeting is optimal for our setting. We therefore vary the LoRA target modules while keeping the training signal and evaluation protocol fixed.

\autoref{fig:target-module-ablation} varies the LoRA target modules across MLP-only, all-linear, and attention-only while keeping the training signal and evaluation protocol fixed. 
The most informative comparison is between MLP-only and all-linear, since all-linear adapts the same MLP modules plus the attention projections. Surprisingly, MLP-only outperforms all-linear in most settings, even though all-linear adapts a broader set of modules and has larger capacity.
For example, in the no-context setting on NarrativeQA, MLP-only reaches 0.207 ROUGE-L versus 0.208 for all-linear; on LongHealth the gap is similarly small (0.231 vs 0.221). 

Attention-only adapters are consistently the weakest, trailing MLP-only by 0.02 to 0.09 ROUGE-L across both datasets. This is consistent with the view that document LoRA serves primarily as factual memory stored in the MLP rather than as a modified attention routing mechanism. Since all-linear adaptation adds attention modules on top of MLP without consistent gains, these results support the MLP-only recipe used in the main experiments.

\section{Conclusion}

We investigate how document-specialized LoRA adapters interact with KV-cache compression as two complementary sources of document memory for question answering. Through a controlled evaluation protocol that progressively evicts document KV states across two datasets and two models, we find three consistent patterns: (1) document LoRA is more helpful at intense KV compression; (2) the base model produces stronger document KV representations, while the adapter is more effective when applied only during answer generation; and (3) QA-style supervision produces substantially stronger adapters than raw-context, continuation, or context-plus-QA objectives. These findings imply that document LoRA should not be viewed as a replacement for long-context encoding, but rather as a complementary parametric memory channel whose value emerges precisely when context-side evidence becomes scarce, suggesting a practical recipe of base-model prefill combined with adapter-augmented decoding under tight KV-cache budgets.

\newpage

\section{Limitations}

Our goal is diagnostic rather than to propose a complete LoRA-augmented serving system. We assume that the correct document adapter is available at inference time, whereas a deployed system would also need an adapter retrieval or routing layer, a policy for composing evidence from multiple documents, and mechanisms for managing adapter storage and loading latency. These requirements do not affect the central comparison in this paper, because all conditions are evaluated with the same document and differ only in whether document information is carried by retained KV states, by the document adapter, or by both. Rather, our results identify the regime in which such systems are most likely to be useful: repeated queries over the same document, private or fixed document collections, or settings where KV-cache budgets force severe context compression. In one-off query settings with ample context budget, standard long-context or retrieval-augmented inference may remain preferable.

\bibliography{custom}

\newpage

\appendix
\label{sec:appendix}

\section{Method Details}
\label{sec::method details}
\subsection{QA Prompt Formats}
\label{app:qa-prompt-formats}
\label{app:training-prompts}

\paragraph{Training.}
The default QA training example is a chat-formatted question-answer pair. It does not include the original document chunk. The user message contains the generated question and asks the model to place the answer in boxed notation:

\begin{lstlisting}[style=promptblockstyle]
<User_token>:
Question: {question}
Answer the question by putting the answer inside the brackets of the "\boxed{}" notation.

<Assistant_token>:
The answer is \boxed{{answer}}.
\end{lstlisting}

During training, the question and prompt prefix are masked, and loss is applied
only to the answer continuation inside the boxed answer:

\begin{lstlisting}[style=promptblockstyle]
>>> loss_mask_start
Question: {question}
Answer the question by putting the answer inside the brackets of the "\boxed{}" notation.
The answer is \boxed{
>>> loss_mask_end
{answer}}.
\end{lstlisting}

\paragraph{Evaluation.}
Evaluation uses the same question-answer template and boxed-answer convention,
but includes the document context before the question. The full evaluation
prompt has the following form:

\begin{lstlisting}[style=promptblockstyle]
<User_token>:
Here is a context:

{document context}

Question: {question}
Answer the question by putting the answer inside
the brackets of the "\boxed{}" notation.

<Assistant_token>:
The answer is \boxed{
\end{lstlisting}

The model begins generation after the fixed answer prefix
"\texttt{The answer is \textbackslash boxed}", so the generated continuation
should complete the boxed answer:

\begin{lstlisting}[style=promptblockstyle]
{answer}}.
\end{lstlisting}

For context-prefill evaluation, this same prompt is split into a document
prefix and a question suffix. The document prefix contains the chat user prefix
and document context, and the question suffix contains the question, answer
instruction, assistant prefix, and fixed boxed-answer prefix. In the no-context
setting, the document context is omitted, and the model receives only the
question and fixed answer prefix.

\paragraph{Raw-context supervision.}
Raw-context training uses the document chunk itself as the training sequence, without a template:

\begin{lstlisting}[style=promptblockstyle]
{raw document chunk text}
\end{lstlisting}

\subsection{Training Supervision Construction}
\label{app:training-supervision-construction}

Synthetic QA data are generated from document chunks before adapter training. In the main QA setting, each document is split into target chunks with a default target length of 64 words and no overlap. For each target chunk, we use fixed dispatch and request four short-answer QA pairs (\texttt{pairs\_per\_call}=4) with one generation call per chunk-task request. The synthesis model is the same model for training the adapter: Qwen3-4B generates the synthetic QA data for Qwen adapters, and Llama-3.1-8B-Instruct generates the synthetic QA data for Llama adapters. During synthesis, neighboring chunks may be supplied as background context, but the prompt instructs the generator to draw questions and answers only from the target passage. Generated outputs are parsed as JSON arrays with \texttt{question} and \texttt{answer} fields; outputs with more than the requested number of items are truncated to the requested count, and failed chunk-task requests are retried.

\begin{lstlisting}[style=promptblockstyle]
Below is a target passage, preceded by surrounding context for
background understanding only.

[Background context]
{surrounding_context}

[Target passage]
{chunk_text}

Generate {pairs_per_call} items whose questions and answers are drawn ONLY from facts stated in the [Target passage] above. Each answer must be a short factual phrase (one word to a few words), for example: a person's name, a number, a date, a location, or another specific fact. Prefer the exact wording used in the passage when possible. Return a JSON array where each element is an object with keys "question" and "answer". Do NOT include any explanation or text outside the JSON array.
\end{lstlisting}

When no surrounding context is used, the generator receives only the passage and the same JSON-format instruction. The background context is used only to help generation; in the default \textsc{Synthetic QA} training format, the original document chunk is not included in the training input.

\begin{table*}[h]
\centering
\small
\resizebox{1\linewidth}{!}{%
\begin{tabular}{llll}
\toprule
Format & Source examples & Training input & Loss tokens \\
\midrule
Synthetic QA & Generated QA pairs & Question + boxed answer & Answer only \\
Raw Context & Document text & Raw full document context & All context tokens \\
Context Continuation & Adjacent chunks & Current chunk $\rightarrow$ next chunk & Next chunk only \\
Context + QA & Generated QA pairs with context & Context + question + boxed answer & Answer only \\
\bottomrule
\end{tabular}%
}
\caption{Training supervision formats used in the format ablation.}
\label{tab:training-supervision-formats}
\end{table*}

\paragraph{Synthetic–evaluation QA overlap} 
We also characterize overlap between synthetic training questions and evaluation questions within the same document or patient record. We compare questions under three increasingly permissive criteria: raw exact match, near-duplicate match, and a broader lexical-similarity match. Exact matches are extremely rare, ranging from 0 to 6 questions across the four source/model settings. Near-duplicate matches are also rare, with at most 16 matched synthetic questions. Even under the broader lexical-similarity criterion, only 13–88 synthetic questions are matched, corresponding to 0.18\%–0.59\% of generated synthetic questions. We do not filter these pairs in the main experiments because the synthetic QA data are generated from document chunks rather than from evaluation annotations; the statistic is reported to make the degree of overlap transparent.

\section{Dataset Statistics}
\label{app:dataset-stats}

Table~\ref{tab:dataset-stats} summarizes the document lengths used in our
main NarrativeQA and LongHealth experiments. We count tokens with the
Llama-3.1-8B-Instruct tokenizer, without adding special tokens, after
deduplicating repeated QA rows by document identifier.

For NarrativeQA, we construct a 20-document subset from the official
NarrativeQA documents and question-answer annotations. We use full story text
rather than summaries or retrieval snippets. Candidate stories are randomly
shuffled with a fixed seed, filtered to keep medium-length full-document
examples, and the first 20 eligible documents are selected. The length filter
only determines document eligibility; selected stories are not truncated. All
available QA pairs for the selected documents are kept, yielding 596 evaluation
examples. For LongHealth, each patient or case record is treated as one
document, giving 20 documents and 400 evaluation examples.

\begin{table}[h]
\centering
\small
\resizebox{\columnwidth}{!}{%
\begin{tabular}{lrrrrr}
\toprule
Dataset & Questions & Docs & Mean & Min & Max \\
\midrule
NarrativeQA & 596 & 20 & 16{,}022 & 11{,}436 & 22{,}974 \\
LongHealth  & 400 & 20 & 11{,}081 & 9{,}418  & 12{,}332 \\
\bottomrule
\end{tabular}%
}
\caption{Document-level context length statistics for the main evaluation subsets, measured with the Llama-3.1-8B-Instruct tokenizer.}
\label{tab:dataset-stats}
\end{table}

Table~\ref{tab:remaining-context-tokens} gives an approximate interpretation of the compression ratios used in our experiments. The values are computed by multiplying the average document length in each dataset by the fraction of context tokens retained after compression. For example, 99.5\% compression keeps only 0.5\% of the document context, corresponding to roughly 80 NarrativeQA tokens or 55 LongHealth tokens on average. These numbers are approximate because the implementation also preserves non-document prompt/template tokens, but they show the scale of the context bottleneck imposed by high compression.

\begin{table}[h]
\centering
\small
\begin{tabular}{lrrrrr}
\toprule
Dataset & 80\% & 90\% & 95\% & 99\% & 99.5\% \\
\midrule
NarrativeQA & 3{,}204 & 1{,}602 & 801 & 160 & 80 \\
LongHealth  & 2{,}216 & 1{,}108 & 554 & 111 & 55 \\
\bottomrule
\end{tabular}
\caption{Approximate remaining document-context tokens after KV-cache compression, computed from the average document lengths in Table~\ref{tab:dataset-stats}.}
\label{tab:remaining-context-tokens}
\end{table}

\section{KV-Cache Compression Implementation}
\label{app:compression-implementation}

We implement KV-cache compression using \texttt{kvpress}~\cite{devoto2025expectedattention} CompactorPress, wrapped with local bookkeeping to support document-context-only compression. In the main generative QA setting, we use a \texttt{context\_prefill} path: the prompt is split into a document prefill prefix and a question suffix. The model first runs a forward pass over the chat/template prefix and document context, producing a KV cache. Compression is applied during this prefill pass by forward hooks on the model's self-attention layers. The question and answer-formatting suffix is then evaluated against the compressed context cache using \texttt{past\_key\_values}. Thus, compression changes the document KV states available to the model, while the question tokens and answer prefix are not themselves compressed.

For each compressed layer, the press scores KV positions and keeps the top-scoring positions according to the requested retention rate. Our default scorer is Compactor, which combines a non-causal attention score with a leverage-score estimate of each key vector's statistical importance. The implementation applies this scoring only during prompt/context prefill; decode steps are skipped, so generated answer tokens are appended to the already-compressed cache rather than triggering new compression decisions. We use \texttt{chunk\_size=256} for the non-causal attention scoring used by Compactor.

The compression ratio $\rho$ is interpreted as the fraction of document-context KV states removed. In the \texttt{context\_prefill} implementation, non-document prompt tokens before the document context are protected, and the requested compression rate is converted into an effective full-prefill rate so that approximately $(1-\rho)$ of the document context remains. For example, $\rho=99.5\%$ keeps roughly $0.5\%$ of the document context tokens, plus a small number of protected prompt/template tokens. The no-compression condition is represented separately and leaves the full document context cache intact.

\begin{figure*}[h]
    \centering
    \includegraphics[width=\linewidth]{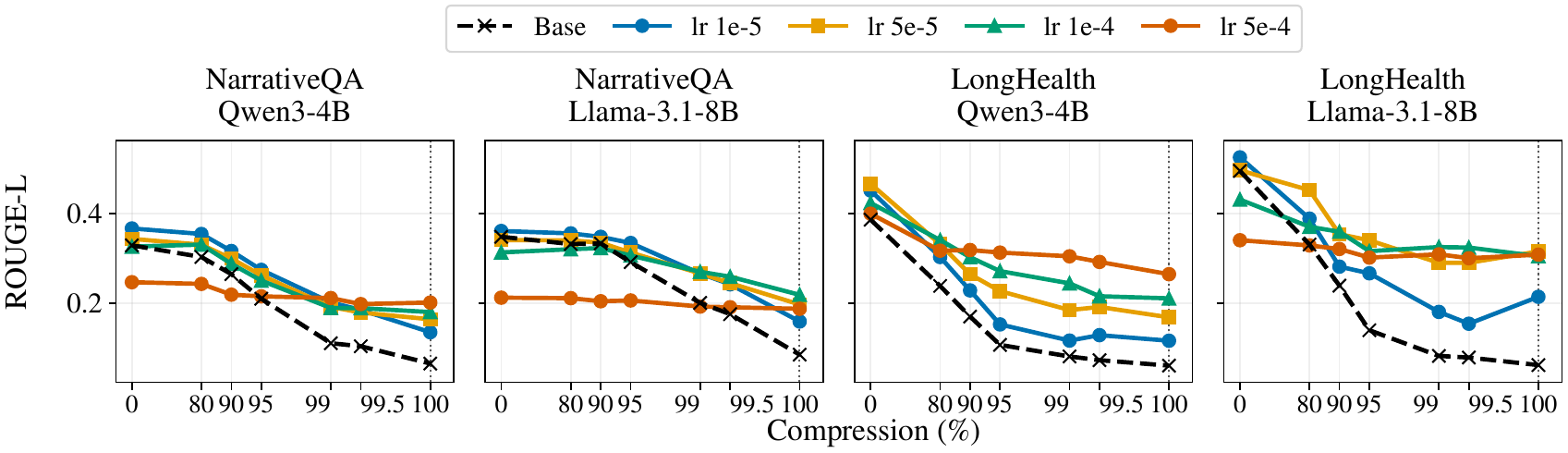}
    \caption{Controlled learning-rate ablation for compressed-generation performance. Each panel plots mean ROUGE-L across the same compression grid used in the main experiments, ending at the 100\% no-context point. Learning-rate choices can change robustness under severe compression even when full-context performance is similar.}
    \label{fig:appendix-learning-rate}
\end{figure*}

\begin{figure*}[h]
    \centering
    \includegraphics[width=\linewidth]{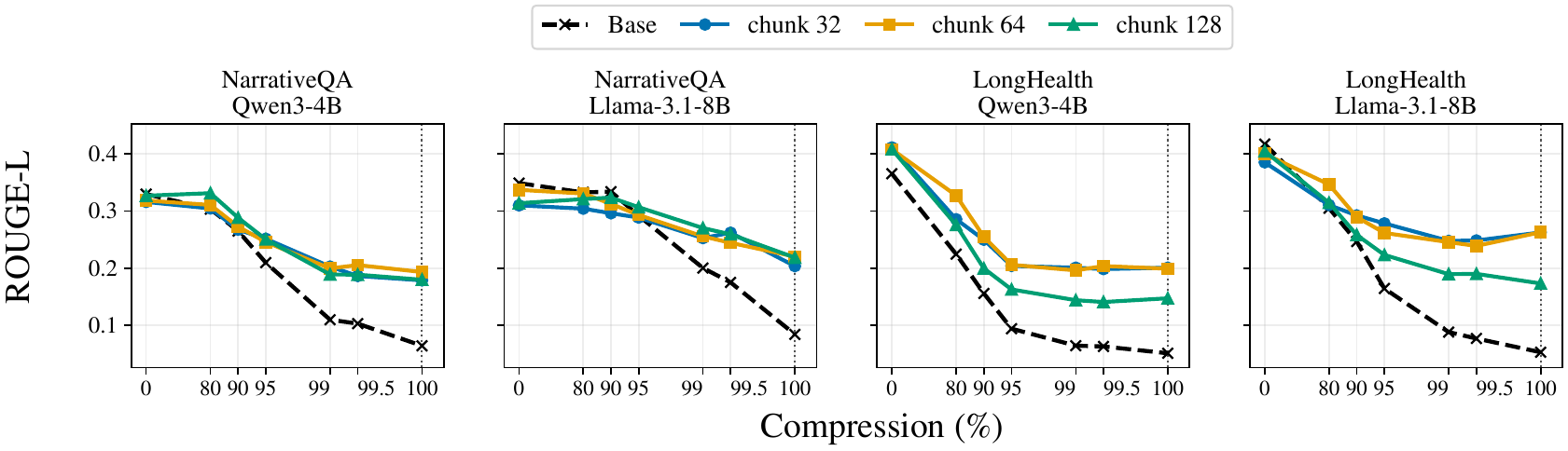}
    \caption{Chunk-size ablation on NarrativeQA and LongHealth. Moderate chunk sizes provide the most stable high-compression behavior: chunk 64 is strongest at the no-context endpoint on NarrativeQA, while chunk 32 and chunk 64 are generally more reliable than the larger-chunk reference on LongHealth.}
    \label{fig:appendix-chunk-size}
\end{figure*}

\section{Additional Training Hyperparameter Controls}
\label{app:training-hyperparameter-controls}

\subsection{Learning-Rate Search}
\label{app:learning-rate-search}

Figure~\ref{fig:appendix-learning-rate} shows the learning-rate sweep used to choose the default adapter training recipe. Across the sweep, $1\times10^{-4}$ provides the most stable tradeoff between full-context performance and high-compression/no-context behavior, so we use it as the default learning rate in the main experiments. This search also shows why tuning only on the full-context point can be misleading: the learning rate that preserves the best uncompressed score is not always the one that gives the most useful adapter behavior when context-side document evidence is heavily compressed or removed.

\subsection{Chunk-Size Search}
\label{app:chunk-size-search}

Figure~\ref{fig:appendix-chunk-size} shows the chunk-size sweep used to choose the default document chunking recipe. We compare chunk sizes 32, 64, and 128 under the same compression grid as the main experiments. On NarrativeQA, chunk sizes 64 and 128 behave similarly in the high-compression region, but chunk 64 is slightly more favorable at the no-context endpoint and avoids moving to a longer training context without clear gain. On LongHealth, the larger chunk-128 setting is noticeably weaker under severe compression, while chunk sizes 32 and 64 are much closer to each other. Since chunk 64 is competitive with chunk 32 on LongHealth and more favorable on NarrativeQA, we use chunk size 64 as the default in the main experiments. Overall, the sweep suggests that moderately sized local document chunks are sufficient for strong compressed-generation behavior, and simply increasing the chunk length does not reliably improve the adapter.

\section{AI Use Disclosure}
We used Claude (Anthropic) and ChatGPT (OpenAI) for writing assistance and editing suggestions. All scientific content, experimental design, and analysis were conducted by the authors.

\end{document}